\documentclass[nohyperref]{article}

\usepackage{microtype}
\usepackage{graphicx}
\usepackage{subfigure}
\usepackage{booktabs} 

\usepackage{hyperref}

\usepackage[accepted]{icml2022}

\usepackage{amsmath}
\usepackage{amssymb}
\usepackage{mathtools}
\usepackage{amsthm}

\usepackage[capitalize,noabbrev]{cleveref}

\theoremstyle{plain}

\theoremstyle{definition}

\theoremstyle{remark}

\usepackage[textsize=tiny]{todonotes}

\icmltitlerunning{Addressing Optimism Bias in Sequence Modeling for Reinforcement Learning}

\begin{document}

\twocolumn[
\icmltitle{Addressing Optimism Bias in Sequence Modeling for Reinforcement Learning}



\icmlsetsymbol{equal}{*}

\begin{icmlauthorlist}
\icmlauthor{Adam Villaflor}{sch}
\icmlauthor{Zhe Huang}{sch}
\icmlauthor{Swapnil Pande}{sch}
\icmlauthor{John Dolan}{sch}
\icmlauthor{Jeff Schneider}{sch}
\end{icmlauthorlist}

\icmlaffiliation{sch}{Carnegie Mellon University}

\icmlcorrespondingauthor{Adam Villaflor}{avillaflor@cmu.edu}
\icmlcorrespondingauthor{Jeff Schnedier}{jeff.schneider@cs.cmu.edu}

\icmlkeywords{Machine Learning, ICML}

\vskip 0.3in
]



\printAffiliationsAndNotice{}  

\begin{abstract}
Impressive results in natural language processing (NLP) based on the Transformer neural network architecture have inspired researchers to explore viewing offline reinforcement learning (RL) as a generic sequence modeling problem. Recent works based on this paradigm have achieved state-of-the-art results in several of the mostly deterministic offline Atari and D4RL benchmarks. However, because these methods jointly model the states and actions as a single sequencing problem, they struggle to disentangle the effects of the policy and world dynamics on the return. Thus, in adversarial or stochastic environments, these methods lead to overly optimistic behavior that can be dangerous in safety-critical systems like autonomous driving. In this work, we propose a method that addresses this optimism bias by explicitly disentangling the policy and world models, which allows us at test time to search for policies that are robust to multiple possible futures in the environment. We demonstrate our method’s superior performance on a variety of autonomous driving tasks in simulation.
\end{abstract}

\section{Introduction}
Recent works have achieved impressive results in a variety of sequence modeling problems from NLP \cite{nlp, nlp2, gpt} to trajectory prediction \cite{scene_transformer, stacked_transformers, urban_transformer, agent_transformer} by leveraging powerful Transformer \cite{transformer} models. Inspired by these numerous successes, several recent works \cite{DT, TT} have explored ways of reformulating sequential-decision making problems in the offline RL framework as a single sequence modeling problem. In particular, these approaches jointly model the states, actions, and rewards as a single data stream with a high-capacity Transformer. These Transformer-based methods are able to outperform the behavioral policy at test time by either conditioning on desired outcomes when picking actions or leveraging the model to search for high-reward trajectories. The main benefit of this paradigm is that it avoids many of the complexities involved in modern model-free and model-based offline RL algorithms.

There are 2 major issues with treating RL as a single sequence modeling problem like the words or characters in a NLP problem. \textbf{(1)} States and actions are fundamentally different concepts. The agent always has complete control over its action sequences, but often has only limited influence on the resulting state transitions. In adversarial or stochastic environments, the same action in the same state can lead to potentially different outcomes, which affects the likelihood or feasibility of achieving a desired result. This leads to the more practical issue \textbf{(2)}, which is that we often need to perform different optimizations over the policy actions and the potential state transitions. Generally, we want to find the action that maximizes reward, but either in expectation over possible future states or with respect to the worst case scenario. Thus, in safety-critical domains or adversarial games we often want to perform a maximum over potential actions and a minimum over possible futures in the environment. Thus in these types of environments, special considerations are needed to ensure effective planning during deployment. \citet{VQ_planning} demonstrates similar issues when deploying MuZero \cite{muzero} with different MCTS \cite{mcts} frameworks in chess. They find that planning with a single-player variant of MuZero that treats the other player as an unknown part of the environment results in a catastrophic drop in performance relative to the traditional two-player adversarial framework.

Most prior works in offline RL have focused on the mainly deterministic D4RL \cite{d4rl} benchmarks and a variety of weakly stochastic Atari \cite{atari} benchmarks. Therefore, there has been limited focus on the difficulties of deploying such methods in largely stochastic domains. In this work, we instead focus specifically on stochastic safety-critical domains, and use autonomous driving as a representative setting. Understanding the stochastic and multi-modal nature of traffic is critical to safe and robust autonomous driving. For example, the same turning sequence for the ego-vehicle could lead to a successful traversal of an intersection or a crash depending on the unknown intentions of the other agents. In this setting, the deployment strategies described in recent sequence modeling approaches for offline RL will lead to overly optimistic behavior because they do not properly disentangle the effects of the policy and world dynamics on the outcome. Specifically, they are biased to believe that the environment will cooperate with them because those sequences are most likely to lead to desired high returns.

In order to address this optimism bias in prior approaches, we develop a method called SeParated Latent Trajectory Transformer (SPLT Transformer), which learns separate generative Transformer models for the policy and dynamics. Because we focus on the autonomous driving domain, we represent each of these models as discrete latent variable Variational Auto-Encoder (VAE) \cite{vae} as inspired by prior work \cite{multi_futures}. By training two separate discrete latent variable VAE models, we can efficiently search over different possible ego behaviors and their interactions with different possible environment responses. We demonstrate how this planning approach allows us during test-time to search for policies that are robust to many possible futures in the environment.

\section{Related Work}
Many recent works have explored leveraging high-capacity sequence models in sequential decision making problems as well as in autonomous driving. Most of the works in the former area have focused on deterministic environments and thus struggle in stochastic and multi-modal problems like autonomous driving. Most of the works applying sequence models to autonomous driving have focused specifically on joint trajectory prediction for all the vehicular and pedestrian agents in the scene. These methods have been effective in capturing the multi-modal stochastic nature of autonomous driving, but often do not consider how these approaches should be leveraged to generate a robust driving policy during deployment. In this work, we develop a method that incorporates ideas from both of these fields in order to learn sequential policy and dynamics models entirely from offline data. Then, we demonstrate how these models can be leveraged at test-time to perform robust planning solely from quantities inferable from an ego-centric perception system.

\subsection{Sequence-Modeling for Offline RL}
Two major works that have explored reformulating the offline RL problem as a single sequence modelling problem are Decision Transformer \cite{DT} and Trajectory Transformer \cite{TT}. In stochastic environments like autonomous driving, both approaches can act optimistically because they fail to account for the uncontrollable factors in the environment that can affect the ability to achieve a specific return.

Decision Transformer is a return-conditioned model-free method that learns a Transformer-based policy that takes in the historical states and actions and a target return, and outputs the action that is most likely to lead to a trajectory sequence that achieves the target return. In stochastic environments, the difficulties come from picking a suitable target return without strong prior knowledge of the testing domain. This is especially problematic because the distribution of possible returns is heavily dependent on the stochastic transitions of the environment. Thus, setting a large target return that is not always feasible can lead to overly aggressive and optimistic behavior, while setting any lower target return could lead to sub-optimal behavior in situations where the environment does unroll favorably.

Trajectory Transformer is a model-based method that trains a Transformer-based trajectory model that can hallucinate potential trajectories in the environment. In stochastic environments, the main concern is how you use this model to properly search for a suitable next action or trajectory-sequence given the uncontrollable stochastic transitions. Naively deploying NLP-style beam search as proposed in the original Trajectory Transformer paper \cite{TT} without accounting for these uncontrollable stochastic transitions will be biased to explore and pick trajectories where the environment just so happens to unroll favorably. This will similarly lead to optimistic behavior, which can be dangerous in safety-critical environments. To address this issue, a different searching procedure needs to be used that can reason about the states and actions separately. One possible avenue for doing this step-wise would be to explore the different variants of MCTS described in \citet{VQ_planning}, but that would lead to an approach that scales exponentially with the search horizon, while our approach only scales linearly with the search horizon.

In order to illuminate the optimism bias in Decision Transformer and Trajectory Transformer, consider a simple discrete and stochastic MDP where there is only 5 states $\{s_0, s_{11}, s_{12}, s_{21}, s_{22}\}\in \mathcal{S}$ and 2 actions $\{a_1, a_2\}\in \mathcal{A}$. The agent always starts in $s_0$. Then, if the agent takes $a_1$ it will stochastically transition from $s_0 \rightarrow s_{11}$ or $s_0 \rightarrow s_{12}$ with equal probability. Similarly, if the agent takes $a_2$ it will stochastically transition from $s_0 \rightarrow s_{21}$ or $s_0 \rightarrow s_{22}$ with equal probability. The reward at each state is: 0 for $s_0$, 10 for $s_{11}$, -10 for $s_{12}$, 6 for $s_{21}$, and 4 for $s_{22}$. The trajectory terminates upon reaching any state besides $s_0$. In this MDP, the expected return for taking $a_1$ is 0 and for taking $a_2$ is 5.
Let's assume that our dataset consists of samples from a uniform random policy.
Now, the optimism bias in Trajectory Transformer comes from the modified beam search they use at test-time. They jointly unroll the next action and resulting state, reward, and return and then filter based on which trajectories have the highest estimated return. In this setting, the trajectory model should properly predict all possible transitions. Then, their beam search will chose to take action $a_1$ because it will lead to the highest return of 10 when it predicts the transition $s_0 \rightarrow s_{11}$. 
However, picking $a_1$ is obviously suboptimal in expectation, even if $a_1$ could lead to the best possible return.
The optimism bias for Decision Transformer comes from picking a fixed target return and their heuristic of setting it to be the highest in the dataset. If we condition on getting a return of 10 (the highest possible), then it will clearly pick action $a_1$ as that is the only action where the probability of achieving the desired outcome is nonzero. Note that this is an issue for any return-conditioned method, not just Decision Transformer.
The inclination for both methods to select $a_1$ because there is a possibility it has the highest return, even though $a_2$ is a better action in expectation and in the worst-case scenario is the specific optimism bias we allude to in this paper.

We validate the optimism bias in these approaches in a toy multi-modal autonomous driving task, and further demonstrate how addressing these issues leads to our method achieving superior performance.

\subsection{Conservatism and Risk-Sensitive RL}
Our work is also related to the fields of risk-sensitive RL and conservatism for offline RL. Most prior works in risk-sensitive RL focus on learning a policy that is optimized not for expected return, but rather some risk-sensitive profile over the distribution of returns \cite{conservative, risk}. Many approaches accomplish this by learning a distributional Q-function that estimates the full distribution of potential returns \cite{iqn} rather than just the expected return. Instead of learning a distributional or risk-sensitive Q-function, we use a learned Transformer world model to hallucinate possible future trajectories and pick a robust behavior that does well in the worst-case predicted future.

Prior works in offline RL have explored different methods of incorporating conservatism in both model-based \cite{mopo, morel} and model-free \cite{cql, iql} RL methods. However, offline RL approaches generally leverage conservatism specifically to discourage the agent from visiting state-actions that are outside the training distribution where the learned models could fail to generalize \cite{offline}. In contrast, in this works we explore using a conservative approach to address the difficulties of planning in stochastic environments in safety-critical domains, like autonomous driving.

\subsection{Trajectory Prediction in Autonomous Driving}
Many recent works in trajectory prediction leverage attention-based models \cite{multi_futures, rnn_atten} and in particular Transformer neural networks \cite{scene_transformer, stacked_transformers, urban_transformer, agent_transformer} to make accurate long-term predictions in complex traffic situations. These methods can learn to attend specifically to the relevant factors in both the target and surrounding vehicles' recent trajectories in order to make predictions that are consistent with the surrounding traffic. Additionally, many trajectory prediction approaches have incorporated VAEs \cite{multi_futures, urban_transformer, agent_transformer} to facilitate covering the different possible modes of future traffic trajectories. However, these approaches have mostly been evaluated on different variations of prediction error on a held out test-set with little focus on the effectiveness of leveraging these models for long-term online planning, like for the CARLA \cite{carla} NoCrash \cite{no_crash} or Leaderboard\footnote{https://leaderboard.carla.org/} benchmarks.

In this work, we also leverage Transformers and VAEs in order to make multi-modal trajectory predictions. However, we focus specially on learning models that can be effectively used for robust search during deployment on an ego vehicle.

\subsection{Learning Behavior for Self-Driving}
Recently, there have been several works that have focused on learning behavioral policies for autonomous driving from offline datasets. Most of this work revolves around performing imitation learning on the behavioral policy of a privileged autopilot agent in simulation \cite{lbc, transfuser, world_rails}.

In our work, we also use a privileged autopilot to collect our offline dataset in simulation. Instead of just learning the behavioral policy, we use these data to learn both a multi-modal policy, and a multi-modal world model that allows us to do efficient and robust search at test-time. Later, we demonstrate how this search leads our method to outperform pure imitation learning.

\section{Preliminaries}
\subsection{Offline Reinforcement Learning}
In RL, we treat the environment as a Markov decision process (MDP)\footnote{While we do evaluate in environments that could be considered POMDPs, we use the MDP formalism in our paper to keep it consist with most prior work in offline RL.} represented as  $(\mathcal{S}, \mathcal{A}, T, \mathcal{R}, \rho_0, \gamma)$, where $\mathcal{S}$ denotes the state space, $\mathcal{A}$ denotes the action space, $T(s' | s, a)$ represents the probabilistic transition dynamics, $\mathcal{R}$ is the reward function, $\rho_0$ is the initial state distribution, and $\gamma \in (0, 1)$ is the discount factor.

For MDPs, a trajectory is a sequence of states and actions
$\tau = \{s_0, a_0, s_1, a_1, \cdots,  s_T, a_T\}$.
Each trajectory has corresponding rewards $r_{\tau}= \{r_0, \cdots,  r_T\}$.
The discounted return for a specific timestep is
\begin{align*}
    R_t = \sum_{i=t}^T \gamma^{t-i} r_{i}
\end{align*}
The goal in RL is to find a policy that maximizes the expected discounted return
\begin{align*}
    \mathbb{E}\left[\sum_{t=0}^T \gamma^{t} r_{t}\right]
\end{align*}


In offline RL, we are a given a fixed dataset $\mathcal{D}_\beta$ consisting of trajectories collected by some behavioral policy $\pi_{\beta}$ in the environment. Without collecting additional data, we must learn a policy that will be effective immediately upon deployment.

\subsection{Transformers}
Transformers \cite{transformer} are a neural network architecture that use several stacks of self-attention blocks to process an arbitrary collection of inputs. By leveraging positional encodings and a causal-attention map, Transformers can be used as sequence models for autoregressive generation. In this work, we mostly use a similar GPT-based \cite{gpt} architecture as DT \cite{DT}. In particular, we use the same linear layer + layer normalization to project the raw inputs into the embedding dimension. We use 4 self-attention blocks with 8 self-attention heads. While our high-level details are more inspired by DT, our code and GPT implementation are based on the publicly available TT \cite{TT} codebase\footnote{https://github.com/JannerM/trajectory-transformer}.

\subsection{Variational Auto-Encoder}
Our method uses conditional variational autoencoders (CVAE) as generative policy and world models that we can use to generate realistic and multi-modal candidate trajectories for test-time search. Thus, in this section we give a general overview of variational autoencoders (VAE). The goal of a VAE model is to generate samples $\hat{d}$ that are within the distribution of a training set $\mathcal{D} = \{d_i\}_{i=1}^N$. Thus the model should be trained to maximize the likelihood of all training points $d_i$. VAEs accomplish this by sampling a latent variable $z$ from some prior $p(z)$ and training a decoder $p_{\theta}(d | z)$ to convert the latent variable into a sample. However, optimizing this model is often intractable and thus \citet{vae} introduces an encoder $q_{\phi}(z | d)$ that can be used to instead optimize the evidence lower bound (ELBO) on the log-likelihood of the data
\begin{align*}
    \log p(d) \geq &E_{z \sim q_{\phi} (z | d)} [\log p_{\theta}(d | z)] \\
    &- \mathcal{D}_{KL}(q_{\phi}(z | d) || p(z))
\end{align*}
The expectation term represents the reconstructions loss, and the KL Divergence term acts as a regularizer that keeps the encoder output close to the prior. Thus with a trained VAE model, we can sample a $z$ from our prior $p(z)$ and pass it to our decoder $p_{\theta}(d | z)$ to generate a sample from the desired distribution.

\subsection{$\beta$-VAE}
When training VAEs with high-capacity models like Transformers or CNNs \cite{cnn}, the decoders often learn to ignore the high-entropy stochastic latent variables. Therefore, \citet{bvae} proposed a modification to the standard ELBO loss and introduced a coefficient $\beta$ to the KL Divergence term. Lowering this hyperparameter $\beta$ below 1 reduces the regularizing effect on the latent variables and thus makes it easier for the models to incorporate the latent variables into their predictions. We find $\beta \in [0.0001, 0.01]$ to be effective for our experiments.

\section{Separated Latent Trajectory Transformer}
\begin{figure*}
\vskip 0.2in
\includegraphics[scale=0.375]{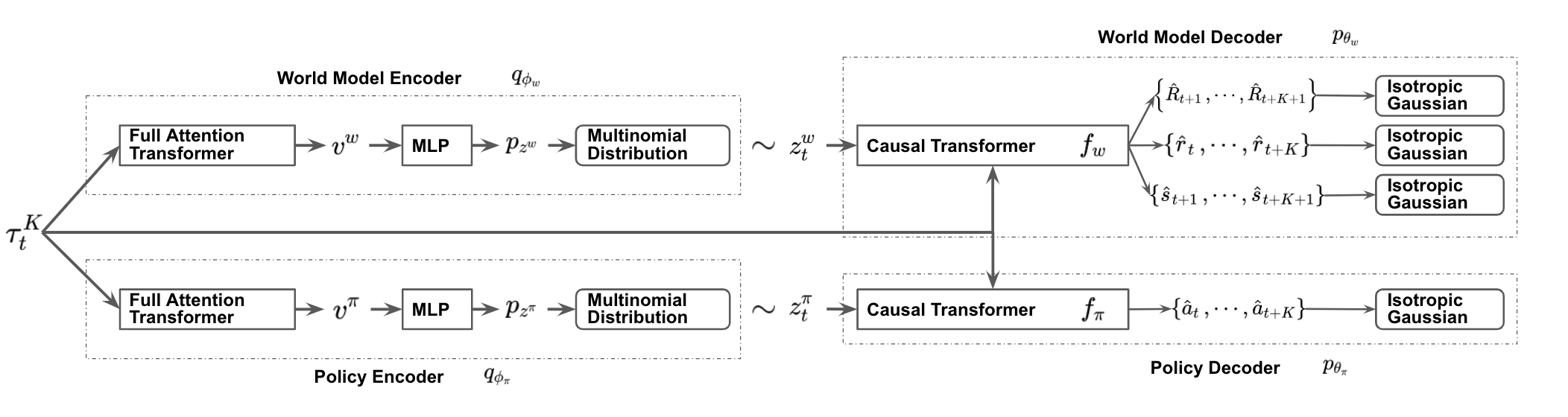}
\centering
\caption{Overview of the SPLT Transformer architecture for generating a reconstruction prediction. The World Model (top) attempts to reconstruct the discounted returns, rewards, and states while the Policy Model (bottom) attempts to reconstruct the action sequence.}
\label{fig:model_architecture}
\vskip -0.2in
\end{figure*}

In this section, we describe how we train 2 separate Transformer-based discrete latent variable VAEs to represent our policy and world models, and how we leverage these models for robust planning at test-time. Figure \ref{fig:model_architecture} contains an overview of the entire SPLT model architecture.

\subsection{Discrete Latent Variable VAE}
In order for our models to be useful for search, they need to \textbf{(1)} be able to produce a good range of candidate behaviors for any given situation for the ego vehicle and \textbf{(2)} cover a majority of the different modes of potential responses from the agents in the environment. Towards this end we train separate Transformer-based VAEs for both the policy and world models. We make the specific design choice for the stochastic latent variables for both models to be discrete and consistent over the entire planning horizon. This allows us to tractably enumerate all possible candidate trajectories without exponential branching, which enables efficient search at test-time. The intuition is that the policy latent variables should correspond to different high-level intentions or policies for the ego-vehicle, like whether to tail another vehicle aggressively or keep your distance. Similarly, the world latent variables should correspond to different possible intentions of the observable vehicles, like whether a vehicle in the opposing lane will go straight or turn through the intersection. Additionally, the world latent variables should capture other events like lights changing or cars suddenly appearing in the sensing range after rounding a corner.

\subsubsection{Encoders}
Both the world encoder $q_{\phi_w}$ and policy encoder $q_{\phi_{\pi}}$ use the same architecture and receive the same trajectory representation. Specifically, they both take in $K$-length trajectory sequences of the form 
\begin{align*}
    \tau_t^K = \{s_t, a_t, s_{t+1}, a_{t+1}, \cdots,  s_{t+K}, a_{t+K}\}
\end{align*}
and output a $n_{w}$ or $n_{\pi}$ dimensional discrete latent variable with each dimension having $c$ possible values
\begin{align*}
   z_t^w &\sim q_{\phi_w}(\cdot | \tau_t^K), &z_t^w &\in \{1, \cdots, c\}^{n_w} \\
   z_t^{\pi} &\sim q_{\phi_{\pi}}(\cdot | \tau_t^K), &z_t^{\pi} &\in \{1, \cdots, c\}^{n_{\pi}}
\end{align*}
The core modules for both encoders are Transformers based on the GPT architecture similar to TT and DT, except we do not perform any masking of the attention. Thus, all elements in the Transformer can fully attend to every other component in the sequence. We take the mean of the Transformer outputs for all the elements in order to coalesce the entire trajectory into a single vector representation $v^{w}$ and $v^{\pi}$. Finally, we pass each of these outputs into a small MLP that outputs the $n_w$ and $n_{\pi}$ independent categorical distributions for $z_t^{w}$ and $z_t^{\pi}$ respectively. Thus, the conditional distributions for $z_t^{w}$ and $z_t^{\pi}$ are represented as $n_w$ and $n_{\pi}$ independent multinomial distributions respectively. We leverage the straight-through gradient estimator\cite{straight_through} as described in \citet{dreamerv2} in order to make the sampling procedure fully differentiable for training.

\subsubsection{Policy Decoder}
The policy decoder uses a similar input trajectory representation
\begin{align*}
    {\tau'}_t^k = \{s_t, a_t, s_{t+1}, a_{t+1}, \cdots,  s_{t+k}\}
\end{align*}
and also takes in the latent variable $z_{\pi}$. The goal of the policy decoder is to estimate
\begin{align*}
    p_{\theta_{\pi}}(a_{t+k} | {\tau'}_t^k; z^{\pi}) \\
    \forall k \in [1, K]
\end{align*}
so that we can predict the most likely next action in the trajectory. We represent this decoder with a causal Transformer model that is very similar to the ones used in Decision Transformer. Beyond excluding the returns-to-go as inputs, the main difference in our model is that we need to incorporate the latent variable $z^{\pi}$. In this work, we incorporate $z^{\pi}$ by first converting it into a single embedding vector, similar to the positional encodings used in other Transformer works, and add it to all the state and action embeddings. We make this design choice instead of inputting the latent variable as another element in the sequence because our method makes it harder for the decoder Transformer to learn to ignore the latent variable, which is a common issue when using high-capacity models in VAEs.

For simplicity, we represent our output distribution as a unit-variance isotropic Gaussian with the mean outputted by our deterministic decoder $f_{\pi}$
\begin{align*}
    p_{\theta_{\pi}}(a_{t+k} | {\tau'}_t^k; z^{\pi}) := \mathcal{N}(f_{\pi}({\tau'}_t^k, z^{\pi}), I)
\end{align*}

\subsubsection{World Model Decoder}

The world model decoder is very similar to the policy decoder, except that its goal is to estimate
\begin{align*}
    p_{\theta_w}(s_{t+k+1} | \tau_t^k; z^{w}) \\
    p_{\theta_w}(r_{t+k} | \tau_t^k; z^{w}) \\
    p_{\theta_w}(R_{t+k+1} | \tau_t^k; z^{w}) \\
    \forall k \in [1, K]
\end{align*}

so that we can predict the most likely next state, reward, and discounted return in the trajectory. The world model decoder is similarly represented with a causal Transformer and incorporates its latent variable $z^w$ in the same manner as the policy decoder. The major difference is that the world model decoder has 3 separate heads to output the 3 different required quantities. The output distributions are similarly represented as unit-variance isotropic Gaussians with the means outputted by the different heads of the deterministic decoder $f_{w}$
\begin{align*}
    p_{\theta_w}(s_{t+k+1} | \tau_t^k; z^{w}) &:= \mathcal{N}(f_{w}^s(\tau_t^k, z^w), I) \\
    p_{\theta_w}(r_{t+k} | \tau_t^k; z^{w}) &:= \mathcal{N}(f_{w}^r(\tau_t^k, z^w), I) \\
    p_{\theta_w}(R_{t+k+1} | \tau_t^k; z^{w}) &:= \mathcal{N}(f_{w}^R(\tau_t^k, z^w), I)
\end{align*}

\subsubsection{Variational Lower Bound}
In order to train our CVAEs, we wish to minimize the standard evidence lower bound (ELBO) on the log-likelihood of the behavioral data for the policy model
\begin{equation}
\begin{split}
    \mathbb{E}_{z_t^{\pi} \sim q_{\phi_{\pi}}} &\left[\sum_{k = 1}^K  \log p_{\theta_{\pi}}(a_{t+k} | {\tau'}_t^k; z_t^{\pi})\right] \\
    &- \mathcal{D}_{KL} (q_{\phi_{\pi}}(z_t^{\pi} | \tau_t^K) || p(z_t^{\pi})) \label{VAE_pi}
\end{split}
\end{equation}
and the world model
\begin{equation}
\begin{split}
    \mathbb{E}_{z_t^{w} \sim q_{\phi_{w}}} &\left[\sum_{k = 1}^K \log p_{\theta_{w}}(s_{t+k+1}, r_{t+k}, R_{t+k+1} | \tau_t^k; z_t^{w})\right] \\
    &- \mathcal{D}_{KL} (q_{\phi_{w}}(z_t^{w} | \tau_t^K) || p(z_t^w)) \label{VAE_w}
\end{split}
\end{equation}
The main difference between our CVAE formulation and the standard formulation is that we use independent discrete uniform distributions as our prior
\begin{align*}
    p(z_{\pi})_i &:= \mathcal{U}\{1, c\} &\forall i \in [1, n_{\pi}] \\
    p(z_{w})_i &:= \mathcal{U}\{1, c\} &\forall i \in [1, n_w]
\end{align*}
because our latent variables are discrete and multidimensional. This prior makes the KL-Divergence terms in equations ($\ref{VAE_pi}$) and ($\ref{VAE_w}$) equivalent to independently maximizing the entropy of each dimension of $z^{\pi}$ and $z^{w}$ respectively. This regularizes the decoders and encourages the VAEs to leverage all available combinations of discrete latent variables. 

Because we use the straight-through gradient estimator, we can differentiate through the multinomial sampling of the latent variables. Thus for both CVAEs, we can jointly train the encoders and decoders end-to-end by directly optimizing objective (\ref{VAE_pi}) for the policy models and objective (\ref{VAE_w}) for the world models.

\subsection{Training}
During training, we sample batches of $K$-length trajectories from our offline dataset. The states and actions of these trajectories are passed into the world and policy encoders in order to generate the corresponding $z^w$ and $z^{\pi}$s. Each $z^{\pi}$ and its corresponding trajectory are passed into the policy decoder, where it predicts all actions in the trajectory through the standard teacher-forcing procedure \cite{teacher_forcing}. Each $z^{w}$ and its corresponding trajectory are passed into the world decoder, where it predicts all the next states, rewards, and discounted returns in the trajectory also through the teacher-forcing procedure. The policy decoder and encoder parameters are updated to minimize equation ($\ref{VAE_pi}$), and the world decoder and encoder parameters are updated to minimize equation ($\ref{VAE_w}$). We train all our models with the Adam optimizer \cite{adam} with a learning rate of $1\mathrm{e}{-4}$ and weight decay $0.1$. Additionally, we normalize all raw values by subtracting the mean and dividing by the standard deviation of the dataset.

\subsection{Planning}

In this section, we describe how we can use our trained conditional policy and world model decoders in order to perform efficient and robust search at test-time.

\subsubsection{Candidate Trajectory Generation}
First, we describe how we can generate a single candidate trajectory given a specific $z^{\pi}$ and $z^w$.

Assume that we are currently at a state $s_t$ and we have stored a history of the last $k$ steps
of the trajectory
\begin{align*}
    {\tau'}_{t-k}^k = \{s_{t-k}, a_{t-k}, s_{t-k+1}, a_{t-k+1}, \cdots,  s_{t}\}
\end{align*}
Our goal is to predict a possible continuation of that trajectory over the planning horizon $h$
\begin{align*}
    \hat{\tau}_{t-k}^{k+h} = \{s_{t-k}, a_{t-k}, \cdots, s_{t}, \hat{a}_{t}, \hat{s}_{t+1}, \cdots,  \hat{s}_{t+h}, \hat{a}_{t+h}\}
\end{align*}
Additionally, we want to estimate the future discounted returns for our candidate trajectory
\begin{align*}
    \hat{R}(\hat{\tau}_t^h) = \sum_{i=0}^{h} \left[ \gamma^{i} \hat{r}_{t+i} \right] + \gamma^{h + 1} \hat{R}_{t+h+1}
\end{align*}

In order to predict these quantities, we alternatively make autoregressive predictions from the policy and world models. Specifically, we alternate between predicting the next action
\begin{align*}
    \hat{a}_{t+i} = f_{\pi}(\{s_{t-k}, a_{t-k}, \cdots,  s_{t}, \hat{a}_{t}, \cdots,  \hat{s}_{t+i}\}, z^{\pi})
\end{align*}
and the next state, reward, and return
\begin{align*}
    \hat{s}_{t+i+1}, &\hat{r}_{t+i}, \hat{R}_{t+i+1} = \\
    &f_{w}(\{s_{t-k}, a_{t-k}, \cdots,  s_{t}, \hat{a}_{t}, \cdots,  \hat{s}_{t+i}, \hat{a}_{t+i}\}, z^{w})
\end{align*}
We repeat this alternating procedure until we reach the horizon length $h$ and compute $\hat{\tau}_t^h$ and its corresponding $\hat{R}(\hat{\tau}_t^h)$.

\subsubsection{Action Selection}
Because we use discrete latent variables, we can enumerate all possible combinations of $z^{\pi}$ and $z^w$. There are $c^{n_{\pi}}$ possible values for $z^{\pi}$ and $c^{n_w}$ possible values for $z^w$, which leads to $c^{n_{\pi}+n_w}$ different possible candidate trajectories. In this work, we found $c=2$, $n_{w} \leq 4$, and $n_{\pi} \leq 4$ to be sufficient for all our explored problems. Thus, we only need to consider a maximum of 256 different combinations of latent variables, which is a standard batch size in many deep learning applications. Therefore, it is easy to run the candidate trajectory generation procedure previously described for all combinations of latent variables on modern GPU hardware.

Without loss of generality we can order the possible values of $z^{\pi}$ and assign each one an index $i \in [1, c^{n_{\pi}}]$. We can do the same for $z^w$ and assign each one an index $j \in [1, c^{n_w}]$. Then, we will label the candidate trajectories that are produced when conditioned on the $i$th $z^{\pi}$ and $j$th $z^w$ as $\hat{\tau}_{ij}$ and its corresponding return $\hat{R}_{ij}$. Then, we select the candidate trajectory that corresponds to
\begin{align*}
    \max_i \min_j \hat{R}_{ij}
\end{align*}
We execute the first action of $\hat{\tau}_{i^*j^*}$ and repeat this procedure at every timestep. The intuition behind this procedure, is that we are trying to pick a policy to follow that will be robust to any realistic possible future in the current environment. Later, we will show how this procedure allows our method to be opportunistic in safe situations and cautious in more dangerous situations.

\section{Experiments}
For all experiments, we compare our SPLT Transformer\footnote{Our code is available at https://github.com/avillaflor/SPLT-transformer} method to Trajectory Transformer (TT), Decision Transformer (DT), and Behavioral Cloning (BC) with a Transformer model. Additionally, on the CARLA benchmarks we compare to Implicit Q-Learning (IQL)\footnote{We use the publicly available rlkit implementation at https://github.com/rail-berkeley/rlkit.} \cite{iql}, which is a state-of-the-art non-recurrent model-free offline RL approach.

\subsection{Illustrative Example}

\begin{table*}[ht]
    \caption{We evaluate all methods with 3 seeds and on 100 different trials in the environment. We report the mean and standard deviation across seeds. DT(m) is DT conditioned on the maximum return in the dataset. DT(e) is DT conditioned on the expected return of the best controller used to collect the dataset. DT(t) is DT with a hand tuned conditional return. TT(a) is TT with more aggressive search parameters. IDM(t) is the best controller from the distribution used to collect the data.}
    \label{tab:toy_results}
    \vskip 0.15in
    \small
    \centering
    \begin{tabular}{| c | c | c | c | c | c | c | c | c | c |} \hline
        \textbf{Metric} & \textbf{SPLT (Ours)} & \textbf{BC} & \textbf{DT(m)} & \textbf{DT(e)} & \textbf{DT(t)} &  \textbf{TT} & \textbf{TT(a)} & \textbf{IDM(t)} \\ \hline
        Return & $78.5 \pm 0.2$ & $58.0 \pm 1.0$ & $28.5 \pm 0.4$ & $68.9 \pm 0.4$ & $73.4 \pm 0.8$ & $58.8 \pm 0.5$ & $67.3 \pm 5.6$ & $78.6$ \\ \hline
        Success (\%) & $100.0 \pm 0.0$ & $100.0 \pm 0.0$ & $50.0 \pm 0.0$ & $100.0 \pm 0.0$ & $100.0 \pm 0.0$ & $100.0 \pm 0.0$ & $91.7 \pm 5.8$ & $100.0$ \\ \hline
    \end{tabular}
    \vskip -0.1in
\end{table*}

We start with a toy autonomous driving problem that we designed to be very simple, but that still demonstrates the dangerous optimism bias in prior Transformer-based approaches. 

In this toy problem, we have an ego vehicle trailing a leading vehicle with both travelling in the same direction on a $1$-D path. Both vehicles are represented using point-mass dynamics, but only the ego vehicle is controllable. Half of the time the leading vehicle will begin hard-braking at the last possible moment in order to stop just before the 70m mark before continuing. The other half of the time the leading vehicle will immediately speed up to the maximum speed and continue for the entire trajectory. The ego vehicle cannot infer beforehand whether the leading vehicle will brake or not, and thus this is a completely stochastic event from the perspective of the ego vehicle.

The observation space is the absolute position and velocity of both the ego and leading vehicle. The action space is just the acceleration for the ego vehicle clipped to $[-1, 1]m/s^2$. The maximum velocity for both vehicles is $10m/s$ and the minimum velocity is $0 m/s$, so the vehicles cannot travel backwards. The ego vehicle is rewarded for the distance traveled at each timestep, but is given a penalty of $-100$ if it crashes. The trajectory ends after $10s$ or if the ego vehicle crashes into the leading vehicle. The ego vehicle is initialized at $0m$ with a random velocity in $[7.5, 10]m/s$. The leading vehicle is initialized randomly within $[10, 20]m$ and with the same velocity as the ego vehicle.

For the offline dataset, we collected $\mathrm{\sim}{100000}$ steps with a distribution of different IDM \cite{idm} controllers that demonstrate a wide range of aggressiveness, and includes some trajectories where the IDM controller is too aggressive and collides with the leading vehicle. We show our results in Table \ref{tab:toy_results}.

For DT, we find that conditioning on the maximum return in the dataset leads to crashes every time the leading vehicle brakes. If we condition on the mean return of the best controller used to collect the dataset, then we get the opposite behavior. The agent does not crash, but also does not take full advantage of the situations where the leading vehicle does not brake, and thus underperforms. While we were able to tune the conditional return in order to get reasonable results, we found the optimal value to be quite arbitrary. Thus, we believe this parameter will be very difficult to tune in more general and complex stochastic environments, where the possible returns would be very hard to estimate beforehand without significant prior knowledge of the specific testing scenario.

For TT, we find that the results depend heavily on the scope of the search used. When we use the default parameters from their codebase for the beam search, we get results very comparable to behavior cloning, which is quite suboptimal. When we reduce the low-probability filtering to allow for more aggressive search, we find that the method sometimes crashes into the leading vehicle because it picks the predicted trajectory where both vehicles will continue at max speed. Similar to DT, we expect tuning the search aggressiveness for TT to be difficult without significant prior knowledge of the intended testing scenario.

For our method, we find that our world VAE is able to predict both possible modes for the leading vehicle's behavior. Additionally, the policy VAE seems to be able to predict a range of different trailing behaviors. Thus, our method is able to properly search for an effective and robust behavior and achieves results comparable to the best controller in the distribution used to collect the data.

\subsection{NoCrash}

Next, we evaluate our method on the CARLA \cite{carla} NoCrash \cite{no_crash} benchmark. For these experiments, we run the $0.9.11$ version of CARLA at $5$fps. We assume access to a global route planner that can generate dense waypoints to our goal, an accurate localization system, and a perception system that can identify the state of any vehicle or traffic light directly in front of us within a limited sensing range. In our experiments, we obtain these ground truth quantities from the CARLA simulator, as commercial self-driving car efforts already have systems to provide these quantities. Thus, we leave implementing such systems as beyond the scope of this project.

The goal in the CARLA NoCrash benchmark is to navigate in a suburban town to a desired goal waypoint from a predetermined start waypoint. The benchmark takes place in the CARLA towns Town01 and Town02 and consists of 25 different routes in each town. 

For our observation we use a low-dimensional vector representation consisting of: \textbf{(1)} the relative heading error to the next target waypoint, \textbf{(2)} the distance from the center of the target lane, \textbf{(3)} the ego vehicle speed, \textbf{(4)} the relative distance to the leading vehicle or the max sensing range if there is no leading vehicle in range, \textbf{(5)} the speed of the leading vehicle or the max speed if there is no leading vehicle in range, \textbf{(6)} the distance to the upcoming red light or the max sensing range if there is no red light in range. There are 2 actions: the steering and the target velocity for a PID controller. The ego car is rewarded for traveling faster and receives a small penalty for deviating from the target lane and a large penalty for crashing or incurring a traffic infraction. We terminate the trajectory whenever the car crashes, incurs an infraction, times out, or reaches the goal.

We collect our offline dataset with autopilot agents with a distribution of different levels of aggressiveness. This aggressiveness corresponds to the parameters of a time-to-collision-based controller that is used to adjust the vehicle's speed in response to any leading vehicles. Additionally, the autopilots use a PID controller for steering, and always immediately brake if they are too close to a red light. We collect $\mathrm{\sim}{300000}$ steps with these autopilots in the Town01 routes in the dense traffic setting. The autopilots are imperfect and have an average success rate of $95.45\%$ in Town01.

\begin{table*}[ht]
    \caption{We evaluate all methods with 3 seeds and on 4 different runs through all 25 routes in the unseen Town02. We report the mean and standard deviation across seeds. DT(m) is Decision Transformer conditioned on the maximum return in the dataset. DT(t) is DT with a hand tuned conditional return. Autopilot(t) is the best controller from the distribution used to collect the data}
    \label{tab:no_crash}
    \vskip 0.15in
    \centering
    \begin{tabular}{| c | c | c | c | c | c | c | c |} \hline
        \textbf{Metric} & \textbf{\begin{tabular}[c]{@{}c@{}}SPLT\\ (Ours)\end{tabular}} & \textbf{BC} & \textbf{TT} & \textbf{DT(m)} & \textbf{DT(t)} & \textbf{IQL} & \textbf{Autopilot(t)} \\ \hline
        Success (\%) & $96.3 \pm 2.1$ & $95.3 \pm 0.4$ & $89.0 \pm 2.8$ & $92.0 \pm 7.1$ & $96.0 \pm 3.3$ & $99.0 \pm 0.8$ & $100.0$ \\ \hline
        Speed (m/s) & $2.71 \pm 0.04$ & $2.51 \pm 0.0$ & $2.67 \pm 0.1$ & $2.72 \pm 0.03$ & $2.72 \pm 0.02$ & $2.84 \pm 0.09$ & $2.75$ \\ \hline
    \end{tabular}
    \vskip -0.1in
\end{table*}

We evaluate all methods by training on this Town01 dataset and then running in the unseen Town02 routes with the dense traffic setting. We depict our results in Table \ref{tab:no_crash}. We find that TT and DT conditioned on the max return have a lower success rate than BC, which we suspect is due to the optimism bias we previously described leading to unnecessary collisions and infractions. When we tune the target return for DT it can outperform BC in terms of average success rate and speed. However, similar to the toy problem, we found the tuned return to be quite arbitrary. Without online evaluation or prior knowledge of the testing domain, it would be quite difficult to estimate the best target return, especially considering that the testing scenarios are different from the training scenarios. Our SPLT method is also able to achieve a higher average success rate and speed compared to BC. We suspect that our positive results are due to our method's planning procedure which avoids the optimism bias of TT's naive beam search.

IQL is a non-recurrent model-free offline RL method. Instead of using model-based planning or a return-conditioned policy, it leverages a conservative Q-learning approach to perform advantage weighted regression\cite{AWR} in order to learn a policy that improves over the data-collection policy.
IQL's superior performance compared to all the offline Transformer-based approaches demonstrates the benefits of leveraging a conservatively trained Q-function. We believe the contributions of our work related to addressing the optimism bias in prior Transformer-based offline RL methods to be mostly orthogonal to IQL. Thus, we leave potential extensions of incorporating an IQL style Q-function into the planning or policy learning procedure as future work. 

\subsection{Leaderboard}

\begin{table*}[ht]
\centering
\caption{We train and evaluate each methods using 3 seeds on the Leaderboard devtest routes. We report the mean and standard deviation across seeds. Total scores are calculated using the official Leaderboard evaluator, which depends on collisions per kilometer (km), infractions per km, and route completion rate. For both Total score and Completion $(\%)$ a larger value is better, while for Collision (/km) and Infraction (/km) a smaller value is better.}
\label{tab:leaderboard-results}
\vskip 0.15in
\centering
\begin{tabular}{|c|c|c|c|c|c|c|c|} \hline
\textbf{Metric} & \textbf{SPLT (Ours)} & \textbf{BC} & \textbf{DT(m)} & \textbf{DT(t)} & \textbf{TT} & \textbf{IQL} & \textbf{Autopilot} \\ \hline
Total score & $68.5 \pm 3.9$ & $51.7 \pm 8.7$ & $63.4 \pm 13.4$ & $70.4 \pm 4.8$ & $55.7 \pm 9.8$ & $66.5 \pm 8.5$ & $70.1$ \\ \hline
Completion (\%) & $94.6 \pm 7.7$ & $95.3 \pm 5.4$ & $96.2 \pm 5.4$ & $97.5 \pm 3.6$ & $66.2 \pm 12.4$ & $100.0 \pm 0.0$ & $100.0$ \\ \hline
Collision (/km) & $1.9 \pm 0.5$ & $4.6 \pm 1.0$ & $1.7 \pm 1.9$ & $1.5 \pm 0.8$ & $3.2 \pm 1.8$ & $2.1 \pm 0.2$ & $1.4$ \\ \hline
Infraction (/km) & $2.3 \pm 1.1$ & $2.4 \pm 1.2$ & $2.7 \pm 0.7$ & $2.4 \pm 0.4$ & $0.0 \pm 0.0$ & $2.1 \pm 1.5$ & $1.8$ \\ \hline
\end{tabular}
\vskip -0.1in
\end{table*}

Next, we evaluate our method on a modified version of the CARLA Leaderboard\footnote{https://leaderboard.carla.org/} benchmark. For these experiments, we run the $0.9.10.1$ version of CARLA at $10$fps. The only additional assumption we make is that our perception system can identify the state of any vehicles or pedestrians directly surrounding us in all directions within a limited sensing range.

The CARLA Leaderboard benchmark is much more comprehensive and requires the agent to perform more involved maneuvers like lane-changing in urban and highway situations. The major difference from our NoCrash setup is that we introduce 8 additional variables to the observation, corresponding to the distance and speed of surrounding vehicles in each of the 4 diagonal directions. We collect $\mathrm{\sim}{1.2}$ million time steps using the autopilot from the Transfuser \cite{transfuser} codebase\footnote{https://github.com/autonomousvision/transfuser} in the CARLA Challenge 2021 training routes for our offline dataset. We evaluate all methods on the officially released devtest routes. We depict our results in Table \ref{tab:leaderboard-results}.

Besides DT with a tuned conditional-return, our method achieves the best overall driving score among all methods. Driving score is a comprehensive indicator for driving quality that accounts for route completion, collisions, and traffic infractions. The increased complexity in the Leaderboard scenarios leads to the world being less predictable and cooperative from the ego-vehicle's perspective. Thus, our SPLT Transformer method, which disentangles the world dynamics and the agent decision-making process, is better equipped to handle this stochastic and uncooperative environment.

We believe that TT in particular underperforms relative to our SPLT method because its naive beam search does not plan appropriately for the range of possibly uncooperative multi-modal outcomes. On the other hand, our method is even able to achieve results on par with the privileged autopilot we used to collect the dataset.

\section{Conclusion}
We presented our SeParated Latent Trajectory Transformer (SPLT Transformer) method, which trains 2 separate policy and world VAE models that can be used at test-time to efficiently perform robust search. We discussed how our approach avoids the optimism bias that other Transformer-based approaches for offline RL struggle with in stochastic settings. Finally, we demonstrated how our method outperforms these baseline approaches on a variety of autonomous driving tasks.

\textbf{Acknowledgments}: This work was supported by the
CMU Argo AI Center for Autonomous Vehicle Research.

\bibliography{paper}
\bibliographystyle{icml2022}

\newpage
\appendix
\onecolumn
\section{Hyperparameters}
For all Transformer-based methods across all experiments, we kept the general Transformer hyperparameters consistent. We used 4 layers of self-attention blocks with 8 heads and an embedding size of 128.

For all Transformer-based methods, we start with the linear warmup rate scheduler from the TT codebase\footnote{https://github.com/JannerM/trajectory-transformer}. For all methods except TT, after the warmup we used a learning rate of $1\mathrm{e}{-4}$ with a constant learning rate schedule. For TT, we used the cosine-annealing schedule from their codebase.

For our SPLT method, the only additional important hyperparameters are $c$, $n_w$,  and $n_{\pi}$ for the latent variables, $\beta$ for the VAE, and $h$ and $k$ for the planning. We generally did a hyperparameter search over $n_w \in [2, 4]$, $n_{\pi} \in [2, 4]$, $\beta \in \{1\mathrm{e}{-4}, 1\mathrm{e}{-3}, 1\mathrm{e}{-2}\}$, $h \in \{5, 10\}$ and $k \in \{2, 5\}$. For, the toy illustrative problem we used $c=2$, $n_w=2$, $n_{\pi}=3$, $\beta = 1\mathrm{e}{-3}$, $h=5$, and $k=5$. For NoCrash, we used $c=2$, $n_w=3$, $n_{\pi}=2$, $\beta = 0.01$, $h=5$, and $k=2$. For Leaderboard, we used $c=2$, $n_w=3$, $n_{\pi} = 2$, $\beta = 0.01$, $h=5$, and $k=2$.

For BC, we used a context $k=5$ for all experiments. For DT, we similarly used a context $k=5$ for all experiments. For TT, we used a context $k=5$ and horizon $h=5$ and the default search parameters from their codebase for all experiments. For IQL, we use the default parameters provided in the RLKit \footnote{https://github.com/rail-berkeley/rlkit} implementation. Namely, we set $\tau=0.7$, $\beta=0.33$ where $\beta$ is an inverse temperature parameter for the advantage term in the policy extraction step.

For DT(t), we search over different scalars $\alpha$ times the max return in the dataset as the desired target return. For the toy problem, we searched over $\alpha \in \{0.8, 0.81, 0.82, 0.83, 0.84, 0.85, 0.86, 0.87, 0.88, 0.89, 0.9, 1.0\}$ and found $\alpha = 0.86$ to get the best results. For NoCrash, we searched over $\alpha \in \{0.3, 0.4, 0.5, 0.6, 0.7, 0.8, 0.9, 1.0\}$ and found $\alpha = 0.6$ to get the best results. For Leaderboard, we searched over $\alpha \in \{0.25, 0.5, 0.75, 1.0, 2.0\}$ and found $\alpha = 0.5$ to get the best results.

\section{Offline D4RL Experiments}
\begin{table*}[ht]
    \caption{For completness, we compare our approach to existing state-of-the-art offline RL methods on the standard D4RL Mujoco locomotion-v2 domain. For BC, MBOP, CQL, DT, and TT we use the results reported from the TT paper. For IQL, we use the results reported from their paper. We report the mean and std for our method over 3 seeds with 10 trajectories for each seed.}
    \label{tab:d4rl_results}
    \vskip 0.15in
    \centering
    \begin{tabular}{ c c c c c c c c c} \hline
        \textbf{Dataset} & Environment & BC & MBOP & CQL & DT & TT & IQL & SPLT(ours) \\ \hline
        Med-Expert & HalfCheetah & $59.9$ & $105.9$ & $91.6$ & $86.8$ & $95.0 \pm 0.2$ & $86.7$ & $91.8 \pm 0.5$  \\
        Med-Expert & Hopper & $79.6$ & $55.1$ & $105.4$ & $107.6$ & $110.0 \pm 2.7$ & $91.5$ & $104.8 \pm 2.6$ \\
        Med-Expert & Walker2d & $36.6$ & $70.2$ & $108.8$ & $108.1$ & $101.9 \pm 6.8$ & $109.6$ & $108.6 \pm 1.1$  \\ \hline
        Medium & HalfCheetah & $43.1$ & $44.6$ & $44.0$ & $42.6$ & $46.9 \pm 0.4$ & $47.4$ & $44.3 \pm 0.7$  \\
        Medium & Hopper & $63.9$ & $48.8$ & $58.5$ & $67.6$ & $61.1 \pm 3.6$ & $66.3$ & $53.4 \pm 6.5$ \\
        Medium & Walker2d & $77.3$ & $41.0$ & $72.5$ & $74.0$ & $79.0 \pm 2.8$ & $78.3$ & $77.9 \pm 0.3$  \\ \hline
        Med-Replay & HalfCheetah & $4.3$ & $42.3$ & $45.5$ & $36.6$ & $41.9 \pm 2.5$ & $44.2$ & $42.7 \pm 0.3$  \\
        Med-Replay & Hopper & $27.6$ & $12.4$ & $95.0$ & $82.7$ & $91.5 \pm 3.6$ & $94.7$ & $75.0 \pm 23.8$ \\
        Med-Replay & Walker2d & $36.9$ & $9.7$ & $77.2$ & $66.6$ & $82.6 \pm 6.9$ & $73.9$ & $57.7 \pm 4.7$  \\ \hline
        & Average & $47.7$ & $47.8$ & $77.6$ & $74.7$ & $78.9$ & $76.9$ & $72.9$
    \end{tabular}
    \vskip -0.1in
\end{table*}

For completeness, we evaluate our method on the D4RL Mujoco locomotion-v2 tasks. In these determinstic tasks, we find that our approach is comparable, but has no obvious competitive advantage over the prior Transformer-based methods as expected. Specifically, we find that our SPLT approach is generally competitive with these approaches on the medium-expert and medium datasets, but underperforms on the hopper and walker2d medium-replay tasks. We believe these results are reasonable given that the medium-replay datasets are the least similar to the autonomous driving data setting, where we assume that the dataset contains a limited number of temporally consistent behaviors.

For all these locomotion tasks we used $c=2$, $n_w=1$, $n_{\pi}=4$, $\beta = 0.01$, $h=10$, and $k=5$

\section{Additional Toy AV Results}

In table \ref{tab:toy_ablate_results}, we show an additional ablation result for our method, where we take a max over world latent variables instead of a min. This corresponds to our planner picking the candidate trajectory based on

\begin{align*}
    \max_i \max_j \hat{R}_{ij}
\end{align*}

Using this planning procedure causes our approach to have a similar optimism bias as the beam search used in TT. We find that using this max planner gets substantially worse results and causes collisions with the leading vehicle. These results demonstrate the importance in our approach of searching for a policy latent variable that is robust to the worst-case scenario in stochastic environments. Additionally, we include results for IQL with default parameters on this toy problem for reference.

\begin{table*}[ht]
    \caption{}
    \label{tab:toy_ablate_results}
    \vskip 0.15in
    \small
    \centering
    \begin{tabular}{| c | c | c | c | c | } \hline
        \textbf{Metric} & \textbf{SPLT (Ours)} & \textbf{SPLT (max planner)} & \textbf{IQL} & \textbf{IDM(t)} \\ \hline
        Return & $78.5 \pm 0.2$ & $46.6 \pm 9.0$ & $68.9 \pm 0.4$ & $78.6$ \\ \hline
        Success (\%) & $100.0 \pm 0.0$ & $69.0 \pm 0.9$ & $100.0 \pm 0.0$ & $100.0$\\ \hline
    \end{tabular}
    \vskip -0.1in
\end{table*}

\section{Planning}
\begin{figure}
\vskip 0.2in
\includegraphics[scale=0.4]{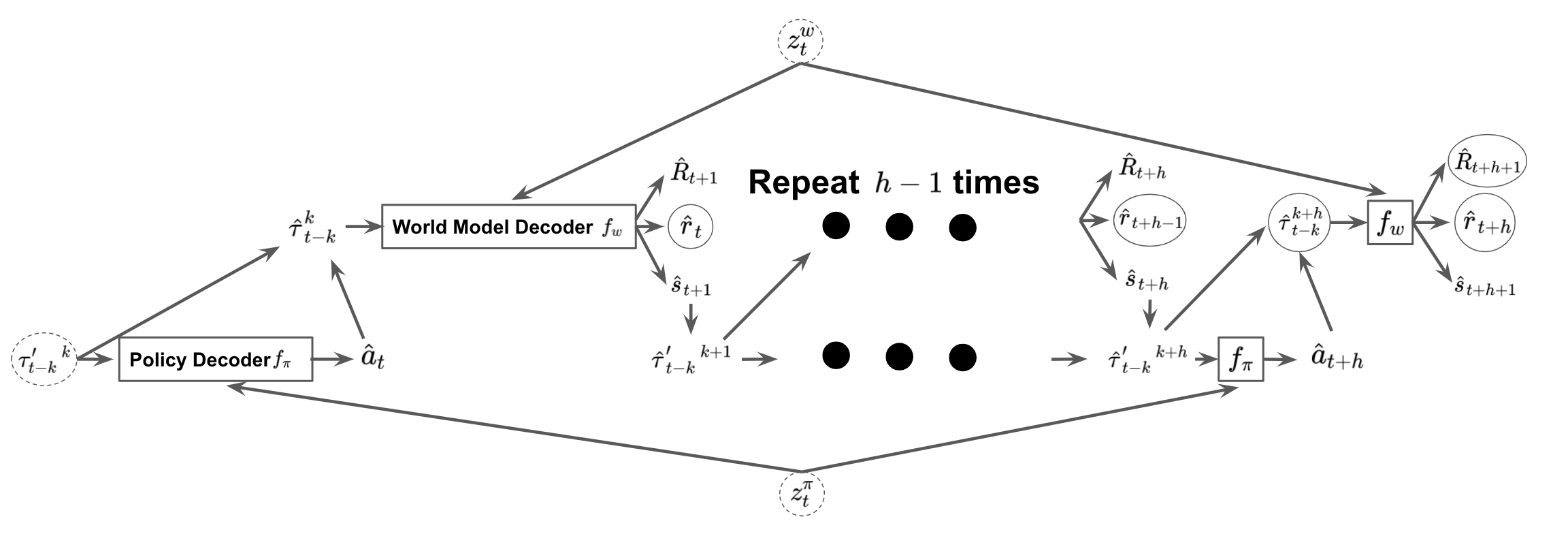}
\centering
\caption{The dashed circled inputs represent the input variable, and the completed circled inputs represent the desired outputs needed for action selection.}
\label{traj_gen}
\vskip -0.2in
\end{figure}

We include an additional diagram (Figure \ref{traj_gen}) to help depict the trajectory generation used in the planning algorithm.

\section{Sensors}
\begin{figure}
\vskip 0.2in
\includegraphics[scale=0.4]{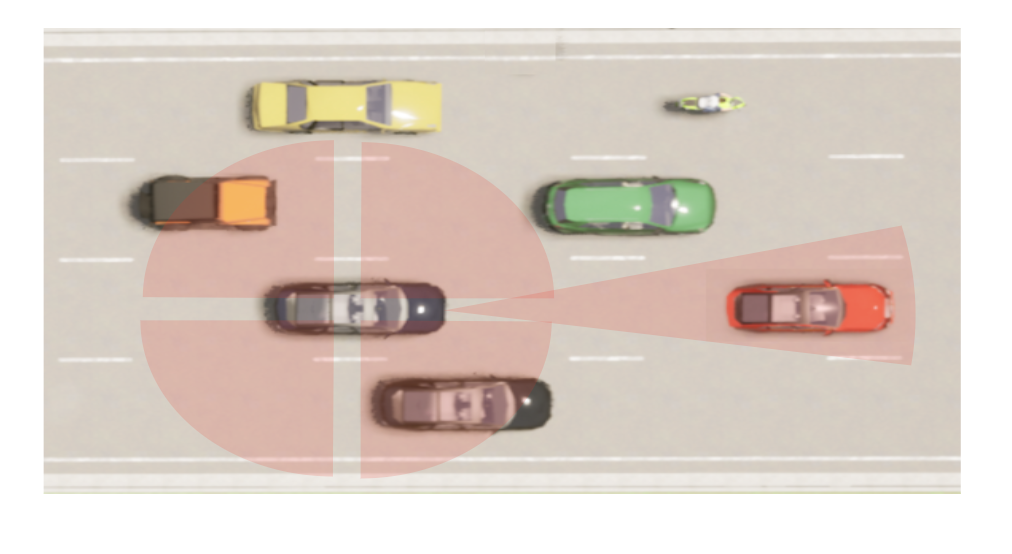}
\centering
\caption{Each shaded red cone represents the sensing region for each of the obstacle detectors. Note that the range is not drawn to scale.}
\label{sens}
\vskip -0.2in
\end{figure}

We include a diagram (Figure $\ref{sens}$) here to depict the obstacle sensor setup we use in the Leaderboard experiments.


\end{document}